\pdfoutput=1
\documentclass[11pt]{article}

\usepackage{acl}

\usepackage{times}
\usepackage{latexsym}

\usepackage[T1]{fontenc}
\usepackage[utf8]{inputenc}

\usepackage{microtype}

\usepackage{amsmath}
\usepackage{multirow}
\usepackage{booktabs}
\usepackage{graphicx}
\usepackage{rotating}

\title{Multilingual unsupervised sequence segmentation transfers to extremely low-resource languages}

\author{C.M. Downey, Shannon Drizin$^*$, Levon Haroutunian$^*$, Shivin Thukral\thanks{~~Accepted to ACL 2022 main conference. Equal contribution from starred authors, sorted by last name. Sincere thanks to: Gina-Anne Levow, Shane Steinert Threlkeld, and Sara Ng for helpful comments and discussion; Francis Tyers for access to the K'iche' data; Manuel Mager for access to the morphologically-segmented validation data} \\
    Department of Linguistics, University of Washington \\
    {\tt \{cmdowney, sdrizin, levonh, shivin7\}@uw.edu} \\}
  
\date{}

\begin{document}
\maketitle
\begin{abstract}
We show that unsupervised sequence-segmentation performance can be transferred to extremely low-resource languages by pre-training a Masked Segmental Language Model \citep{downey_masked_2021} multilingually. Further, we show that this transfer can be achieved by training over a collection of low-resource languages that are typologically similar (but phylogenetically unrelated) to the target language. In our experiments, we transfer from a collection of 10 Indigenous American languages (AmericasNLP, \citealp{mager_findings_2021}) to K'iche', a Mayan language. We compare our multilingual model to a monolingual (from-scratch) baseline, as well as a model pre-trained on Quechua only. We show that the multilingual pre-trained approach yields consistent segmentation quality across target dataset sizes, exceeding the monolingual baseline in 6/10 experimental settings. Our model yields especially strong results at small target sizes, including a zero-shot performance of 20.6 F1. These results have promising implications for low-resource NLP pipelines involving human-like linguistic units, such as the \textit{sparse transcription} framework proposed by \citet{bird_sparse_2020}.
\end{abstract}

\section{Introduction}
Unsupervised sequence segmentation (at the word, morpheme, and phone level) has long been an area of interest in languages without whitespace-delimited orthography \citep[e.g. Chinese,][]{uchiumi_inducing_2015, sun_unsupervised_2018}, morphologically complex languages without rule-based morphological analyzers \citep{creutz_unsupervised_2002}, and automatically phone-transcribed speech data \citep{goldwater_bayesian_2009, lane_computational_2021}, respectively. It has been particularly important for lower-resource languages in which there is little or no gold-standard data on which to train supervised models \citep{joshi_state_2020}.

In modern neural end-to-end systems, unsupervised segmentation is usually performed via information-theoretic algorithms such as BPE \citep{sennrich_neural_2016} and SentencePiece \citep{kudo_sentencepiece_2018}. However, the segmentations they produce are largely non-sensical to humans \citep{park_morphology_2021}. The motivating tasks listed above instead require unsupervised approaches that correlate more closely with human judgements of the boundaries of linguistic units. For example, in a human-in-the-loop framework such as the \textit{sparse transcription} proposed by \citet{bird_sparse_2020}, lexical items are automatically proposed to native speakers for confirmation, and it is important that these candidates be (close to) sensical, recognizable pieces of language.

In this paper, we investigate the utility of recent models that have been developed to conduct unsupervised surface morpheme segmentation as a byproduct of a language modeling objective \citep[e.g.][see Section~\ref{related_work}]{kawakami_learning_2019, downey_masked_2021}. The key idea is that recent breakthroughs in crosslingual language modeling and transfer learning \cite[\textit{inter alia}]{conneau_cross-lingual_2019, artetxe_cross-lingual_2020} can be leveraged to facilitate transferring unsupervised segmentation performance to a new target language, using these types of language models.

Specifically, we investigate the effectiveness of multilingual pre-training in a Masked Segmental Language Model \citep{downey_masked_2021} when applied to a low-resource target. We pre-train our model on the ten Indigenous languages of the 2021 AmericasNLP shared task dataset \citep{mager_findings_2021}, and apply it to another low-resource, Indigenous, and morphologically complex language of Central America: K'iche' (quc), which at least phylogenetically is unrelated to the pre-training languages \citep{campbell_meso-america_1986}.

We hypothesize that multilingual pre-training on similar, possibly contact-related languages, will outperform both a monolingual baseline trained from scratch and a model pre-trained on a single language (Quechua) with the same amount of pre-training data. We also expect that the pre-trained models will perform increasingly better than the monolingual baseline the smaller the target corpus is.

Indeed, our experiments show that a pre-trained multilingual model provides stable performance across all dataset sizes and far exceeds the monolingual baseline at low-to-medium target sizes. We additionally show that the multilingual model achieves a zero-shot segmentation performance of 20.6 F1 on the K'iche' data, where the monolingual baseline yields a score of zero. These results suggest that transferring from a multilingual model can greatly assist unsupervised segmentation in very low-resource languages, even those that are morphologically rich. The results also provide evidence for the idea that transfer from multilingual models works at a more moderate scale than is typical for recent crosslingual models (3.15 million parameters for our models).

In the following section, we overview work relating to unsupervised segmentation, crosslingual pre-training, and transfer-learning (Section~\ref{related_work}). We then introduce the multilingual data used in our experiments, and the additional pre-processing we performed to prepare the data for pre-training (Section~\ref{data}). Next we provide a brief overview of the type of Segmental Language Model used in our experiments, as well as our multilingual pre-training process (Section~\ref{model}). After this, we describe our experimental process applying the pre-trained and from-scratch models to varying target data sizes (Section~\ref{experiments}). Finally, we discuss the results of our experiments and their significance for low-resource pipelines, both within unsupervised segmentation and for other NLP tasks more generally (Sections \ref{results} and \ref{analysis}).

\section{Related Work}\label{related_work}
Work related to the present study largely falls either into the field of (unsupervised) word segmentation, or the field(s) of crosslingual language modeling and transfer learning. To our knowledge, we are the first to propose a multilingual model for unsupervised word/morpheme-segmentation.

\paragraph{Unsupervised Segmentation}
Current state-of-the-art unsupervised segmentation has largely been achieved with Bayesian models such as Hierarchical Dirichlet Processes \citep{teh_hierarchical_2006, goldwater_bayesian_2009} and Nested Pitman-Yor \citep{mochihashi_bayesian_2009, uchiumi_inducing_2015}. Adaptor Grammars \citep{johnson_improving_2009} have been successful as well. Models such as \textit{Morfessor} \citep{creutz_unsupervised_2002}, which are based on Minimal Description Length \citep{rissanen_stochastic_1989} are also widely used for unsupervised morphology.

As \citet{kawakami_learning_2019} note, most of these models have weak language modeling ability, being unable to take into account much other than the immediate local context of the sequence. Another line of techniques has focused on models that are both strong language models and good for sequence segmentation. Many are in some way based on Connectionist Temporal Classification \citep{graves_connectionist_2006}, and include Sleep-WAke Networks \citep{wang_sequence_2017}, Segmental RNNs \citep{kong_segmental_2016}, and Segmental Language Models \citep{sun_unsupervised_2018, kawakami_learning_2019, wang_unsupervised_2021, downey_masked_2021}. In this work, we conduct experiments using the Masked Segmental Language Model of \citet{downey_masked_2021}, due to its good performance and scalability, the latter usually regarded as an obligatory feature of multilingual models \citep[\textit{inter alia}]{conneau_unsupervised_2020, xue_mt5_2021}.

\paragraph{Crosslingual and Transfer Learning}
Crosslingual modeling and training has been an especially active area of research following the introduction of language-general encoder-decoders in Neural Machine Translation, offering the possibility of zero-shot translation (i.e. translation for language pairs not seen during training; \citealp{ha_toward_2016, johnson_googles_2017}).

The arrival of crosslingual language model pre-training \citep[XLM,][]{conneau_cross-lingual_2019} further demonstrates that large models pre-trained on multiple languages yield state-of-the-art performance across an abundance of multilingual tasks including zero-shot text classification (e.g. XNLI, \citealp{conneau_xnli_2018}), and that pre-trained transformer encoders provide great initializations for MT systems and language models in very low-resource languages.

Since XLM, numerous studies have attempted to single out which components of crosslingual training contribute to transferability from one language to another \citep[e.g.][]{conneau_emerging_2020}. Others have questioned the importance of multilingual training, and have instead proposed that even monolingual pre-training can provide effective transfer to new languages \citep{artetxe_cross-lingual_2020}. Though some like \citet{lin_choosing_2019} have tried to systematically study which aspects of pre-training languages/corpora enable effective transfer, in practice the choice is often driven by availability of data and other ad-hoc factors.

Currently, large crosslingual successors to XLM such as XLM-R \citep{conneau_unsupervised_2020}, MASS \citep{song_mass_2019}, mBART \citep{liu_multilingual_2020}, and mT5 \citep{xue_mt5_2021} have achieved major success, and are the starting point for a large portion of multilingual NLP systems. These models all rely on an enormous amount of parameters and pre-training data, the bulk of which comes from very high-resource languages. In contrast, in this paper we assess whether multilingual pre-training on a suite of very low-resource languages, which combine to yield a moderate amount of unlabeled data, can provide good transfer to similar languages which are also very low-resource.

\section{Data and Pre-processing}\label{data}
We draw data from three main datasets. We use the AmericasNLP 2021 open task dataset \citep{mager_findings_2021} to pre-train our multilingual models. The multilingual dataset from \citet{kann_fortification_2018} serves as segmentation validation data for our pre-training process in these languages. Finally, data from \citet{tyers_corpus_2021} is used as the training set for our experiments transferring to K'iche', and \citet{richardson_morphological_2021} provides the validation and test data for these experiments.

\paragraph{AmericasNLP 2021}
The AmericasNLP data consists of train and validation files for ten low-resource Indigenous languages of Central and South America: Asháninka (cni), Aymara (aym), Bribri (bzd), Guaraní (gug), Hñähñu (oto), Nahuatl (nah), Quechua (quy), Rarámuri (tar), Shipibo Konibo (shp), and Wixarika (hch). For each language, AmericasNLP also includes parallel Spanish sets, which we do not use. The data was originally curated for the AmericasNLP 2021 shared task on low-resource Machine Translation. \citep{mager_findings_2021}.\footnote{\url{https://github.com/AmericasNLP/americasnlp2021}}

We augment the Asháninka and Shipibo-Konibo training sets with additional available monolingual data from \citet{bustamante_no_2020},\footnote{\url{https://github.com/iapucp/multilingual-data-peru}} which is linked in the official AmericasNLP repository. We add both the training and validation data from this corpus to the \textit{training} set of our splits.

To pre-process for a multilingual language modeling setting, we first remove lines that contain urls, copyright boilerplate, or that contain no alphabetic characters. We also split lines that are longer than 2000 characters into sentences/clauses where evident. Because we use the Nahuatl and Wixarika data from \citet{kann_fortification_2018} as validation data, we remove any overlapping lines from the AmericasNLP set. We create a combined train file as the concatenation of the training data from each of the ten languages, as well as a combined validation file likewise.

\begin{figure}[ht]
    \begin{center}
    \includegraphics[width=0.45\textwidth]{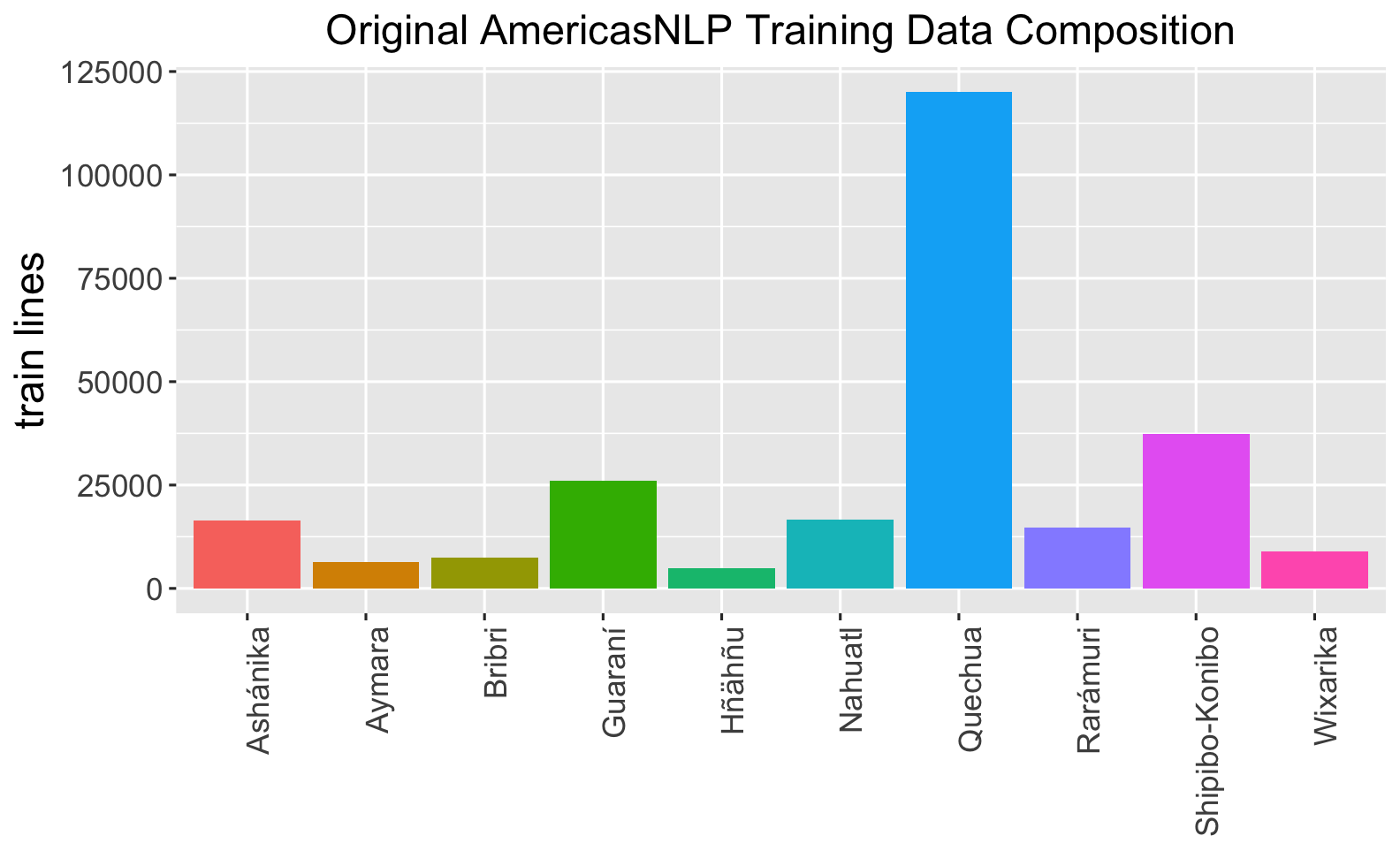}
    \caption{Original (imbalanced) language composition of the AmericasNLP training set}
    \label{anlp_comp_orig}
    \end{center}
\end{figure}

\begin{figure}[ht]
    \begin{center}
    \includegraphics[width=0.45\textwidth]{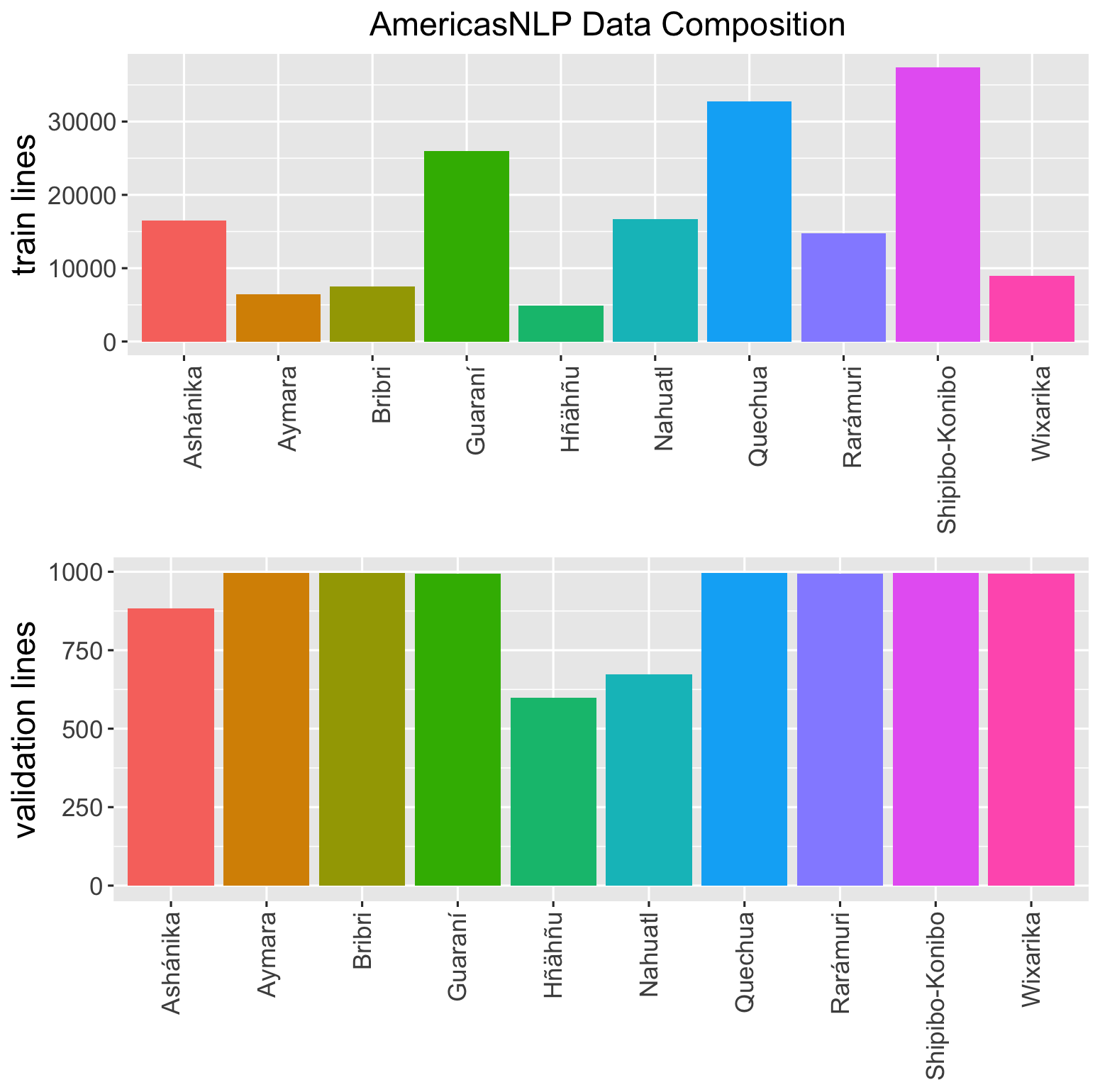}
    \caption{Final language composition of our AmericasNLP splits after downsampling Quechua}
    \label{anlp_comp}
    \end{center}
\end{figure}

Because the original ratio of Quechua training data is so high compared to all other languages (Figure~\ref{anlp_comp_orig}), we downsample it to 2$^{15}$ examples, the closest order of magnitude to the next-largest training set. A plot of the balanced (final) composition of our AmericasNLP train and validation sets is seen in Figure~\ref{anlp_comp}. 

To compare the effect of multilingual and monolingual pre-training, we also pre-train a model on Quechua alone, since it has by far the most data (Figure~\ref{anlp_comp_orig}). However, the full Quechua training set has about 50k fewer lines than our balanced AmericasNLP set (Figure~\ref{anlp_comp}). To create a fair comparison between multilingual and monolingual pre-training, we additionally create a downsampled version of the AmericasNLP set of equal size to the Quechua data (120,145 lines). The detailed composition of our data is available in Appendix~\ref{app:anlp}.

\paragraph{Kann et al (2018)}
The data from \citet{kann_fortification_2018}, originally curated for a segmentation task on polysynthetic low-resource languages, contains morphologically segmented sentences for Nahuatl and Wixarika. We use these examples as validation data for segmentation quality during the pre-training process. We clean this data in the same manner as the AmericasNLP sets.

\paragraph{K'iche' data}
The K'iche' data used in our study was curated for \citet{tyers_corpus_2021}. The raw (non-gold-segmented) data, used as the training set in our transfer experiments, comes from a section of this data web-scraped by the Crúbadán project \citep{scannell_crubadan_2007}. This data is relatively noisy, so we clean it by removing lines with urls or lines where more than half of the characters are non-alphabetic. We also remove duplicate lines. The final data consists of 47,729 examples and is used as our full-size training set for K'iche'. Our experiments involve testing transfer at different resource levels, so we also create smaller training sets by downsampling the original to lower orders of magnitude.

For evaluating segmentation performance on K'iche', we use the segmented sentences from \citet{richardson_morphological_2021},\footnote{\url{https://github.com/ftyers/global-classroom}} which were created for a shared task on morphological segmentation. These segmentations were created by a hand-crafted FST, then manually disambiguated. Because gold-segmented sentences are so rare, we concatenate the original train/validation/test splits and then split them in half into final validation and test sets.

\section{Model and Pre-training}\label{model}

This section gives an overview of the Masked Segmental Language Model (MSLM), introduced in \citet{downey_masked_2021}, along with a description of our pre-training procedure.

\begin{figure}[ht]
    \begin{center}
    \includegraphics[width=0.45\textwidth]{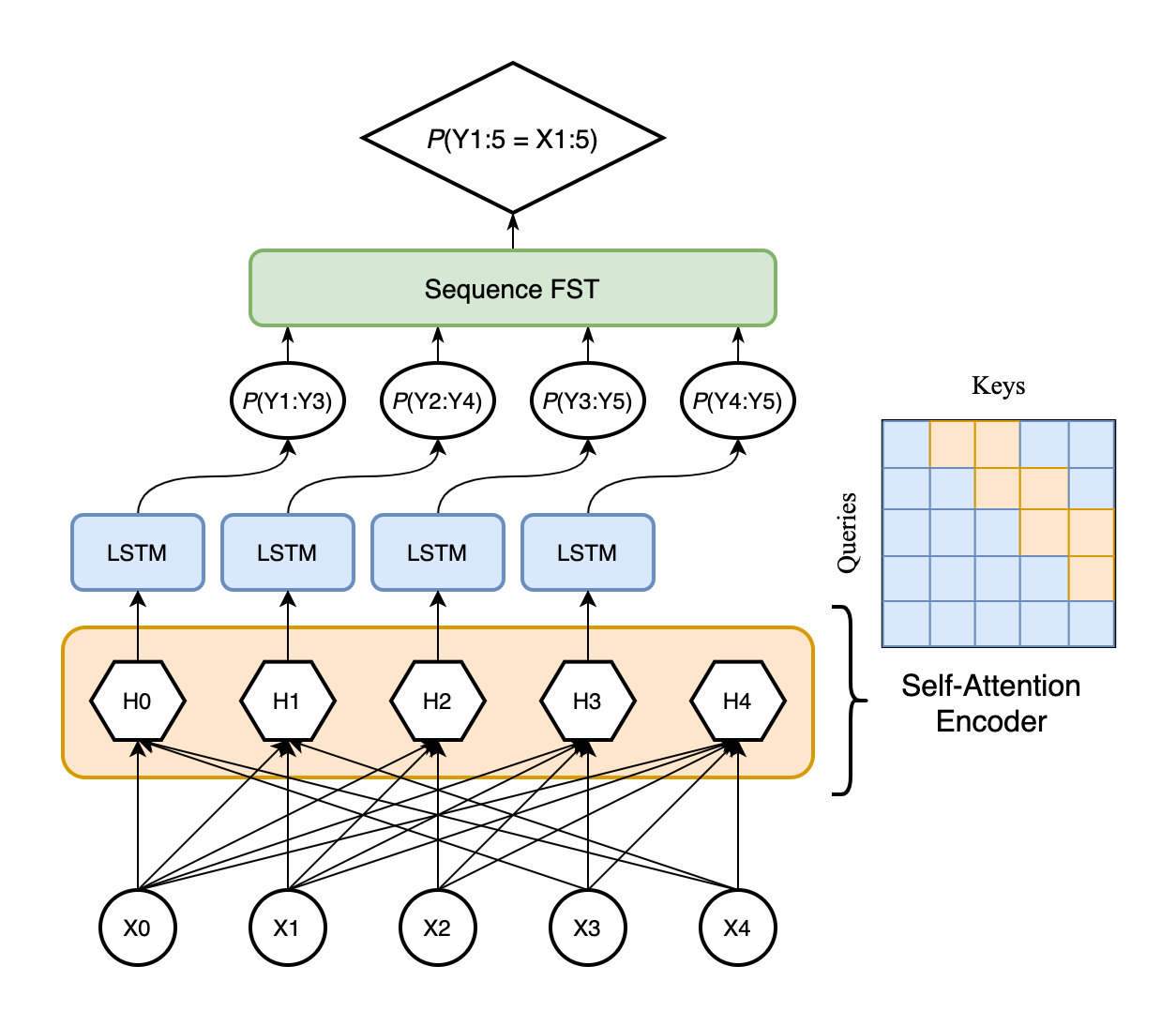}
    \caption{Masked Segmental Language model (left) and Segmental Attention Mask (right). (Figure 3 in \citealp{downey_masked_2021})}
    \label{mslm}
    \end{center}
\end{figure}

\paragraph{MSLMs}
An MSLM is a variant of a Segmental Language Model (SLM) \citep{sun_unsupervised_2018, kawakami_learning_2019, wang_unsupervised_2021}, which takes as input a sequence of characters \textbf{x} and outputs a probability distribution for a sequence of segments \textbf{y} such that the concatenation of \textbf{y} is equivalent to \textbf{x}: $\pi (\textbf{y}) = \textbf{x} $. An MSLM is composed of a Segmental Transformer Encoder and an LSTM-based Segment Decoder \citep{downey_masked_2021}. See Figure~\ref{mslm}.

The MSLM training objective is based on the prediction of masked-out spans. During a forward pass, the encoder generates an encoding for every position in \textbf{x}, for a segment up to $k$ symbols long; the encoding at position $i-1$ corresponds to every possible segment that starts at position $i$. Therefore, the encoding approximates
\[
p(\textbf{x}_{i:i+1}, \textbf{x}_{i:i+2}, ..., \textbf{x}_{i:i+k} | \textbf{x}_{<i}, \textbf{x}_{\ge i+k})
\]

To ensure that the encodings are generated based only on the portions of \textbf{x} that are outside of the predicted span, the encoder uses a Segmental Attention Mask \citep{downey_masked_2021} to mask out tokens inside the segment. Figure~\ref{mslm} shows an example of such a mask with $k = 2$.

Finally, the Segment Decoder of an SLM determines the probability of the $j^{th}$ character of the segment of \textbf{y} that begins at index $i$, $\textbf{y}^i_j$, using the encoded context:
\[
p(\textbf{y}^i_j | \textbf{y}^i_{0:j}, \textbf{x}_{<i}, \textbf{x}_{\ge i+k}) = Decoder(h^i_{j-i}, y^i_{j-1})
\]

The output of the decoder is not conditional on the determination of other segment boundaries. The probability of \textbf{y} is modeled as the marginal probability over all possible segmentations of \textbf{x}. Because directly marginalizing is computationally intractable, the marginal is computed using dynamic programming over a forward-pass lattice. The maximum-probability segmentation is determined by Viterbi decoding. The training objective optimizes language-modeling performance, which is measured in Bits Per Character (bpc).

\paragraph{Pre-training Procedure}
In our experiments, we test the transferability of multilingual and monolingual pre-trained MSLMs. The multilingual models are trained on the AmericasNLP 2021 data (see Section~\ref{data}). Since SLMs operate on plain text, we can train the model directly on the multilingual concatenation of this data, and evaluate it by its language modeling performance on the concatenated validation data. As mentioned in Section~\ref{data}, we create two versions of the multilingual pre-trained model: one trained on the full AmericasNLP set ($\sim$172k lines) and the other trained on the downsampled set, which is the same size as the Quechua training set ($\sim$120k lines). We designate these models \textsc{Multi-pt$_{full}$} and \textsc{Multi-pt$_{down}$}, respectively. Our pre-trained monolingual model is trained on the full Quechua set (\textsc{Quechua-pt}).

Each model is an MSLM with four encoder layers, hidden size 256, feedforward size 512, and four attention heads. Character embeddings are initialized using Word2Vec \citep{mikolov_efficient_2013} over the training data. The maximum segment size is set to 10. The best model is chosen as the one that minimizes the Bits Per Character (bpc) loss on the validation set. For further pre-training details, see Appendix~\ref{app:hypers}.

\begin{figure}[ht]
    \begin{center}
    \includegraphics[width=0.45\textwidth]{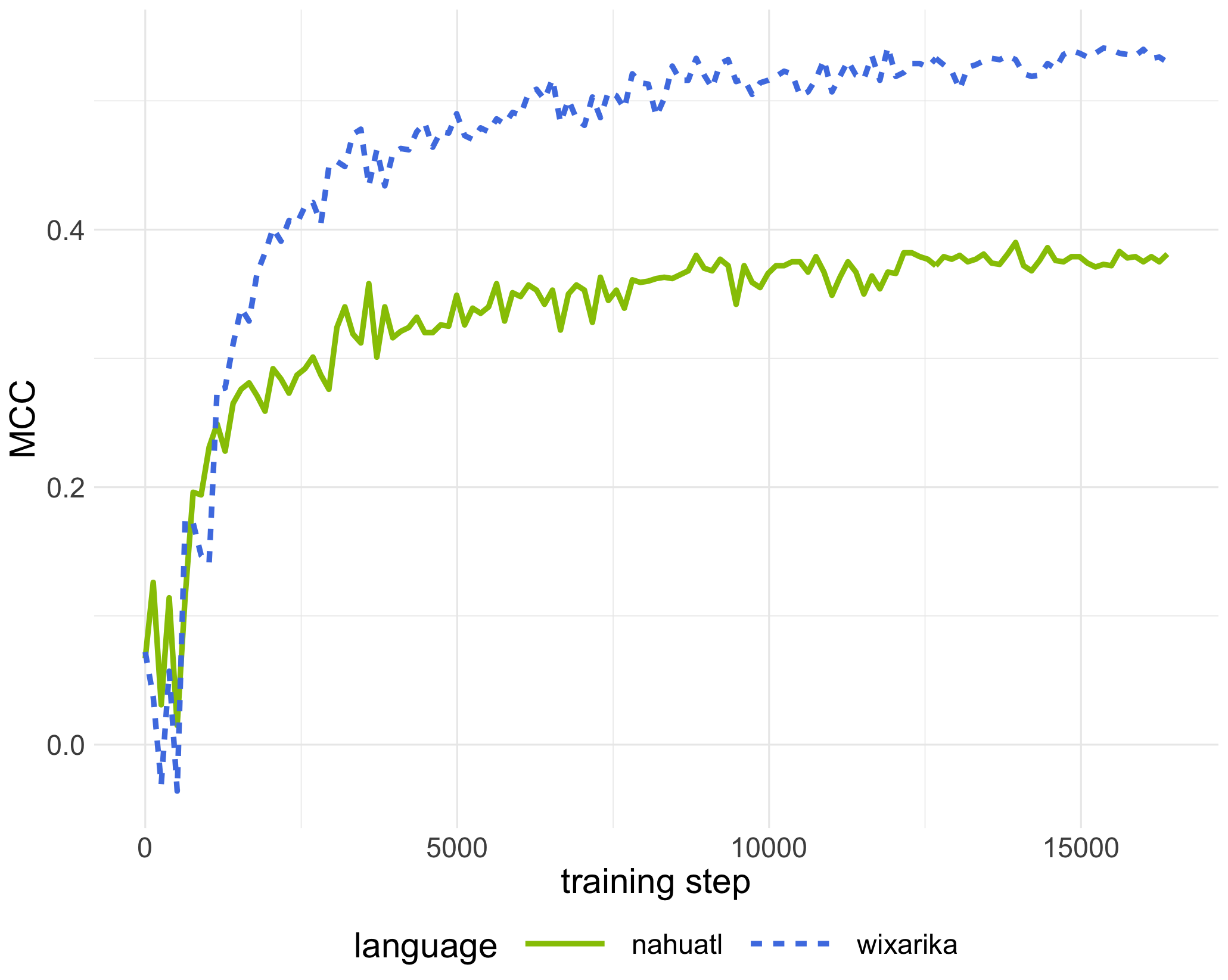}
    \caption{Plot of segmentation quality for Nahuatl and Wixarika during multilingual pre-training (measured by Matthews Correlation Coefficient with gold segmentation)}
    \label{anlp_pretraining_curve}
    \end{center}
\end{figure}

To evaluate the effect of pre-training on the segmentation quality for languages within the pre-training set, we also log MCC between the model output and gold-segmented secondary validation sets available in Nahuatl and Wixarika \citep[see Section~\ref{data}]{kann_fortification_2018}. Figure~\ref{anlp_pretraining_curve} shows the unsupervised segmentation quality for Nahuatl and Wixarika almost monotonically increases during pre-training (\textsc{Multi-pt$_{full}$}).

\section{Experiments}\label{experiments}
We evaluate whether multilingual pre-training facilitates effective low-resource transfer learning for unsupervised segmentation. To do this, we pre-train SLMs on one or all of the AmericasNLP 2021 languages \citep{mager_findings_2021} and transfer it to a new target language: K'iche' \citep{tyers_corpus_2021}. K'iche' is a morphologically rich Mayan language with several classes of inflectional prefixes and suffixes \citep{txchajchal_batz_gramatica_1996}. An example sentence can be found in Table~\ref{kiche_example}, which also shows our model's input and target output format.

As a baseline, we train a monolingual K'iche' model from scratch. We evaluate performance with respect to the size of the target training set, simulating varying degrees of low-resource setting. To do this, we downsample the K'iche' training set to 8 smaller sizes, for 9 total: \{256, 512, ... 2$^{15}$, 47.7k (full)\}. For each size, we both train a monolingual baseline and fine-tune the pre-trained models we describe in Section~\ref{model}.\footnote{All of the data and software required to run these experiments can be found at \url{https://github.com/cmdowney88/XLSLM}}

\begin{table*}[ht]
    \centering
    \begin{tabular}{ll}
    \toprule
    Orthography &  kinch'aw ruk' le nunan \\
    Linguistic Segmentation & k-in-ch'aw r-uk' le nu-nan \\
    Translation & ``I speak with my mother'' \\
    Model Input & \texttt{kinch'awruk'lenunan} \\
    Target Output & \texttt{k in ch'aw r uk' le nu nan} \\
    \bottomrule
    \end{tabular}
    \caption{Example K'iche' sentence from \citet{tyers_corpus_2021}. This sentence consists of multiple words, some of which consist of multiple morphemes. The model receives the sentence as an unsegmented stream of characters and the target output is a sequence of morphemes (word and morpheme boundaries are treated the same, since the former is a subtype of the latter)}
    \label{kiche_example}
\end{table*}

\paragraph{Architecture and Modeling}
All models are Masked Segmental Language Models (MSLMs) with the architecture described in Section~\ref{model}. The only difference is that the baseline model is initialized with a character vocabulary \textit{only} covering the particular K'iche' training set (size-specific). The character vocabulary of the K'iche' data is a subset of the AmericasNLP vocabulary, so we are able to transfer the multilingual models without changing the embedding and output layers. The Quechua vocabulary is \textit{not} a superset of the K'iche', so we add the missing characters to the Quechua model's embedding block \textit{before} pre-training (these are randomly initialized). The character embeddings for the baseline are initialized using Word2Vec \citep{mikolov_efficient_2013} on the training set (again, size-specific). 

\paragraph{Evaluation Metrics}
SLMs can be trained in either a fully unsupervised or ``lightly'' supervised manner \citep{downey_masked_2021}. In the former case, only the language modeling loss (Bits Per Character, bpc) is used to pick parameters and checkpoints. In the latter, the segmentation quality on gold-segmented validation data can be considered. Though our validation set is gold-segmented, we pick the best parameters and checkpoints based on bpc only, simulating the unsupervised case. However, to monitor the change in segmentation quality during training, we also use Matthews Correlation Coefficient (MCC). This measure frames segmentation as a character-wise binary classification task (i.e. boundary vs. no boundary), and measures correlation with the gold segmentation.

To make our results comparable with the wider word-segmentation literature, we use the scoring script from the SIGHAN Segmentation Bakeoff \citep{emerson_second_2005} for our final segmentation F1. For each model and target size, we choose the best checkpoint (by bpc), apply the model to the combined validation and test set, and use the SIGHAN script to score the output.

For comparison to the Chinese Word-Segmentation and speech literature, any whitespace segmentation in the validation/test data is discarded before it is fed to the model. However, SLMs can also be trained to treat spaces like any other character, and thus could be able to take advantage of existing segmentation in the input. We leave this for future work.

\paragraph{Parameters and Trials}
For our training procedure (both training the baseline from scratch and fine-tuning the pre-trained models) we tune hyperparameters on three of the nine dataset sizes (256, 2048, and full) and choose the optimal parameters by bpc. For each of the other sizes, we directly apply the chosen parameters from the tuned dataset of the closest size (on a log scale). We tune over five learning rates and three encoder dropout values. As in pre-training, we set the maximum segment length to 10. For more details on our training procedure, see Appendix~\ref{app:hypers}.

\section{Results}\label{results}
The results of our K'iche' transfer experiments at various target sizes can be found in Table~\ref{tab:results}. In general, the (full) pre-trained multilingual model (\textsc{Multi-pt$_{full}$}) demonstrates good performance across dataset sizes, with the lowest segmentation performance (20.6 F1) being in the zero-shot case and the highest (40.7) achieved on 2$^{14}$ examples. The monolingual baseline outperforms \textsc{Multi-pt$_{full}$} at the two largest target sizes, as well as at size 4096 (achieving the best overall F1 of 44.8), but performs very poorly under 2048 examples, and has no zero-shot ability (unsurprisingly, since it is a random initialization).

Interestingly, other than in the zero-shot case, \textsc{Quechua-pt} and the comparable \textsc{Multi-pt$_{down}$} perform very similarly to each other. However, the zero-shot transferability of \textsc{Multi-pt$_{down}$} is almost twice that of the model trained on Quechua only. \textsc{Multi-pt$_{full}$} exceeds both \textsc{Multi-pt$_{down}$} and \textsc{Quechua-pt} by a wide margin in every setting. Finally, all models show increasing performance until about size 4096, after which more target examples don't provide a large increase in segmentation quality.

\begin{table*}[ht]
    \centering
    \begin{tabular}{lcccccccccc}
        \toprule
        \multirow{2}{*}{Model} & \multicolumn{10}{c}{Target Language Segmentation F1} \\
        \cmidrule{2-11}
        {} & 0 & 256$^*$ & 512 & 1024 & 2048$^*$ & 4096 & 8192 & 2$^{14}$ & 2$^{15}$ & 47,729 (full)$^*$ \\
        \midrule
        \textsc{Multi-pt$_{full}$} & \textbf{20.6} & \textbf{34.0} & \textbf{37.4} & \textbf{37.4} & 38.2 & 40.5 & \textbf{38.6} & \textbf{40.7} & 38.9 & 38.2 \\
        \textsc{Multi-pt$_{down}$} & 15.0 & 25.1 & 25.7 & 29.3 & 32.5 & 33.2 & 33.3 & 31.5 & 33.6 & 31.9 \\
        \textsc{Quechua-pt} & 7.6 & 29.9 & 31.0 & 30.4 & 30.7 & 31.0 & 29.9 & 33.6 & 31.8 & 33.3 \\
        \textsc{Monolingual} & 0.002 & 4.0 & 3.3 & 10.3 & \textbf{39.2}$^*$ & \textbf{44.8} & 29.4 & 39.5 & \textbf{44.1} & \textbf{43.2} \\
        \bottomrule
    \end{tabular}
    \caption{Segmentation quality on the combined validation and test set for each model, at each target training set size. Star indicates size at which hyperparameter tuning is conducted. For tuned sizes, showing only the performance of the model with the best bpc. *See Table \ref{tab:results_parameter_variation}: the best baseline trial achieved slightly better performance than \textsc{Multi-pt$_{full}$}, but the former is far more sensitive to variation due to hyperparameters at this size}
    \label{tab:results}
\end{table*}

\begin{table*}[ht]
    \centering
    \begin{tabular}{lccc}
        \toprule
        \multirow{2}{*}{Model} & \multicolumn{3}{c}{Target Language Segmentation F1} \\
        \cmidrule{2-4}
        {} & 256$^*$ & 2048$^*$ & 47,729 (full)$^*$ \\
        \midrule
        \textsc{Multi-pt$_{full}$} & \textbf{34.2 $\pm$ 0.6} (1.8\%) & \textbf{38.1 $\pm$ 0.4} (1.0\%) & 39.4 $\pm$ 1.1 (2.8\%) \\
        \textsc{Multi-pt$_{down}$} & 25.7 $\pm$ 0.6 (2.3\%) & 30.5 $\pm$ 2.3 (7.5\%) & 31.7 $\pm$ 0.6 (1.9\%) \\
        \textsc{Quechua-pt} & 30.1 $\pm$ 0.2 (0.7\%) & 31.4 $\pm$ 0.6 (1.9\%) & 32.7 $\pm$ 0.7 (2.1\%) \\
        \textsc{Monolingual} & 4.2 $\pm$ 0.5 (11.9\%) & 36.5 $\pm$ 6.8 (18.6\%) & \textbf{44.7 $\pm$ 2.0} (4.5\%) \\
        \bottomrule
    \end{tabular}
    \caption{Variation of segmentation quality across the best four hyperparameter combinations for a single size (by bpc; mean $\pm$ standard deviation (stdev $\div$ mean); models ranked by mean minus stdev)}
    \label{tab:results_parameter_variation}
\end{table*}

\paragraph{Interpretation}
These results show that \textsc{Multi-pt$_{full}$} provides consistent performance across target sizes as small as 512 examples. Even for size 256, there is only a 9\% (relative) drop in quality from the next-largest size. Further, the pre-trained model's zero-shot performance is impressive given the baseline is effectively 0 F1.

On the other hand, the performance of the monolingual baseline at larger sizes seems to suggest that given enough target data, it is better to train a model devoted to the target language only. This is consistent with previous results \citep{wu_are_2020, conneau_unsupervised_2020}. However, it should also be noted that \textsc{Multi-pt$_{full}$} never trails the baseline by more than 5.2 F1.

One less-intuitive result is the dip in the baseline's performance at sizes 8192 and 2$^{14}$. We believe this discrepancy may be partly explainable by sensitivity to hyperparameters in the baseline. Though the best baseline trial at size 2048 exceeds \textsc{Multi-pt$_{full}$} by a small margin, the baseline shows large variation in performance across the top-four hyperparameter settings at this size, where \textsc{Multi-pt$_{full}$} actually performs better on average and much more consistently (Table~\ref{tab:results_parameter_variation}). We thus believe the dip in performance for the baseline at sizes 8192 and 2$^{14}$ may be due to an inability to extrapolate hyperparameters from other experimental settings.

\section{Analysis and Discussion}\label{analysis}
\paragraph{Standing of Hypotheses}
Within the framework of unsupervised segmentation, these results provide strong evidence that relevant linguistic patterns can be learned over a collection of low-resource languages, and then transferred to a new language without much (or any) target training data. Further, it is shown that the target language need not be (phylogenetically) related to any of the pre-training languages, even though details of morphological structure are ultimately language-specific.

The hypothesis that multilingual pre-training yields increasing advantage over a from-scratch baseline at smaller target sizes is also strongly supported. This result is consistent with related work showing this to be a key advantage of the multilingual approach \citep{wu_are_2020}.

The hypothesis that multilingual pre-training also yields better performance than monolingual pre-training given the same amount of data seems to receive mixed support from our experiments. On one hand, the comparable multilingual model has a clear advantage over the Quechua model in the zero-shot setting, and outperforms the latter in 5/10 settings more generally. However, because the Quechua data lacks several frequent K'iche' characters (and these embeddings remain randomly initialized), it is unclear how much of this advantage comes from the multilingual training \textit{per-se}. Instead, the advantage may be due to the multilingual model's full coverage of the target vocabulary--- an advantage which may disappear at larger target sizes. Further analysis of this hypothesis will require additional investigation.

\paragraph{Significance}
The above results, especially the strong zero-shot transferability of segmentation performance, suggest that the type of language model used here learns some abstract linguistic pattern(s) that are generalizable across languages, and even to new ones. It is possible that these generalizations could take the form of abstract stem/affix or word-order patterns, corresponding roughly to the lengths and order of morphosyntactic units. Because MSLMs operate on the character level (and in these languages orthographic characters mostly correspond to phones), it is also possible the model could recognize syllable structure in the data (the ordering of consonants and vowels in human languages is relatively constrained), and learn to segment on syllable boundaries.

It is also helpful to remember that we select the training suite and target language to have some characteristics in common that may help facilitate transfer. The AmericasNLP languages are almost all morphologically rich, with many considered polysynthetic \citep{mager_findings_2021}, a feature that K'iche' shares \citep{suarez_mesoamerican_1983}. Further, all of the languages, including K'iche', are spoken in countries where either Spanish or Portuguese is the official language, and have very likely had close contact with these Iberian languages and borrowed lexical items. Finally, the target language family (Mayan) has also been shown to have close historical contact with the families of several of the AmericasNLP set (Nahuatl, Rarámuri, Wixarika, Hñähñu), forming a Linguistic Area or \textit{Sprachbund} \citep{campbell_meso-america_1986}.

It is possible that one or several of these shared characteristics facilitates the strong transfer shown here, in both our multilingual and monolingual pre-trained models. However, our current study does not conclusively show this to be the case. \citet{lin_choosing_2019} show that factors like linguistic similarity and geographic contact are often not as important for transfer success as non-linguistic features such as the raw size of the source dataset. Indeed, the fact that our Quechua pre-trained model performs similarly to the comparable multilingual model (at least at larger target sizes) suggests that the benefit to using \textsc{Multi-pt$_{full}$} could be interpreted as a combined advantage of pre-training data size and target vocabulary coverage.

The nuanced question of whether multilingual pre-training \textit{itself} enables better transfer than monolingual pre-training requires more study. However, taking a more pragmatic point of view, multilingual training can be seen as a methodology to 1) acquire more data than is available from any one language and 2) ensure broader vocabulary overlap with the target language. Our character-based model is of course different from more common word- or subword-based approaches, but with these too, attaining pre-trained embeddings that cover a novel target language is an important step in cross-lingual transfer \cite[\textit{inter alia}]{garcia_towards_2021, conneau_unsupervised_2020, artetxe_cross-lingual_2020}

\paragraph{Future Work}
We believe some future studies would shed light on the nuances of segmentation transfer-learning. First, pre-training either multilingually or monolingually on languages that are \textit{not} linguistically similar to the target language could help isolate the advantage given by pre-training on \textit{any} language data (vs. similar language data).

Second, we have noted that monolingual pre-training on a language that does not have near-full vocabulary coverage of the target language leaves some embeddings randomly initialized, yielding worse performance at small target sizes. Pre-training a model on a single language that happens to have near-complete vocabulary coverage of the target could give a better view of whether multilingual training intrinsically yields advantages, or whether monolingual training is disadvantaged mainly due to this lack of vocabulary coverage.

Finally, because none of the present authors have any training in the K'iche' language, we are unable to perform a linguistically-informed error analysis of our model's output (e.g. examining the types of words and morphemes which are erroneously (un)segmented, rather than calculating an overall precision and recall for the predicted and true morpheme boundaries, as we do in this study). However, we make all of our model outputs available in our public repository, so that future work may provide a more nuanced analysis of the types of errors unsupervised segmentation models are prone to make.

\section{Conclusion}\label{conclusion}
This study has shown that unsupervised sequence segmentation ability can be transferred via multilingual pre-training to a novel target language with little or no target data. The target language also need not be from the same family as a pre-training language for successful transfer. While training a monolingual model from scratch on large amounts of target data results in good segmentation quality, our experiments show that pre-trained models, especially multilingual ones, far exceed the baseline at small target sizes ($\leq$1024), and seem to be much more robust to hyperparameter variation at medium sizes (2048, 8192, 2$^{14}$).

One finding that may have broader implications is that pre-training can be conducted over a set of low-resource languages with some typological or geographic connection to the target, rather than over a crosslingual suite centered around high-resource languages like English and other European languages. Most modern crosslingual models have huge numbers of parameters (XLM has 570 million, mT5 has up to 13 billion, \citealp{xue_mt5_2021}), and are trained on enormous amounts of data, usually bolstered by hundreds of gigabytes in the highest-resource languages \citep{conneau_unsupervised_2020}.

In contrast, our results suggest that effective transfer may be possible at smaller scales, by combining the data of low-resource languages and training moderately-sized, more targeted pre-trained multilingual models (our model has 3.15 million parameters). Of course, this study can only support this possibility within the unsupervised segmentation task, so future work will be needed to investigate whether transfer to and from low-resource languages can be extended to other tasks.

\bibliography{XLSLM}
\bibliographystyle{acl_natbib}

\appendix
\clearpage
\section{AmericasNLP Datasets}\label{app:anlp}
\paragraph{Composition}
The detailed composition of our preparation of the AmericasNLP 2021 training and validation sets can be found in Tables \ref{anlp_train_table} and \ref{anlp_dev_table} respectively. \texttt{train\_1.mono.cni}, \texttt{train\_2.mono.cni}, \texttt{train\_1.mono.shp}, and \texttt{train\_2.mono.shp} are the additional monolingual sources for Asháninka and Shipibo-Konibo obtained from \citet{bustamante_no_2020}. \texttt{train\_downsampled.quy} is the version of the Quechua training set downsampled to 2$^{15}$ lines to be more balanced with the other languages. \texttt{train.anlp} is the concatenation of the training set of every language before Quechua downsampling, and \texttt{train\_balanced.anlp} is the version after Quechua downsampling. \texttt{train\_downsampled.anlp} is the version of our multilingual set downsampled to be the same size as \texttt{train.quy}. \textsc{Multi-pt$_{full}$} is pre-trained on \texttt{train\_balanced.anlp}, \textsc{Multi-pt$_{down}$} is pre-trained on \texttt{train\_downsampled.anlp}, and \textsc{Quechua-pt} is pre-trained on \texttt{train.quy}.

\paragraph{Citations}
A more detailed description of the sources and citations for the AmericasNLP set can be found in the original shared task paper \citep{mager_findings_2021}. Here, we attempt to give a brief listing of the proper citations.

All of the validation data originates from AmericasNLI \citep{ebrahimi_americasnli_2021} which is a translation of the Spanish XNLI set \citep{conneau_xnli_2018} into the 10 languages of the AmericasNLP 2021 open task.

The training data for each of the languages comes from a variety of different sources. The \textbf{Asháninka} training data is sourced from \citet{ortega_overcoming_2020, cushimariano_romano_naantsipeta_2008, mihas_naani_2011} and consists of stories, educational texts, and environmental laws. The \textbf{Aymara} training data consists mainly of news text from the GlobalVoices corpus \citep{prokopidis_parallel_2016} as available through OPUS \citep{tiedemann_parallel_2012}. The \textbf{Bribri} training data is from six sources \citep{feldman_neural_2020, margery_diccionario_2005, jara_murillo_gramatica_2018, constenla_curso_2004, jara_murillo_se_2013, jara_murillo_i_2018, flores_solorzano_corpus_2017} ranging from dictionaries and textbooks to story books. The \textbf{Guaraní} training data consists of blogs and web news sources collected by \citet{chiruzzo_development_2020}. The \textbf{Nahuatl} training data comes from the Axolotl parallel corpus \citep{gutierrez-vasques_axolotl_2016}. The \textbf{Quechua} training data was created from the JW300 Corpus \citep{agic_jw300_2019}, including Jehovah's Witnesses text and dictionary entries collected by \citet{huarcaya_taquiri_traduccion_2020}. The \textbf{Rarámuri} training data consists of phrases from the Rarámuri dictionary \citep{brambila_diccionario_1976}. The \textbf{Shipibo-Konibo} training data consists of translations of a subset of the Tatoeba dataset \citep{montoya_continuous_2019}, translations from bilingual education books \citep{galarreta_corpus_2017}, and dictionary entries \citep{loriot_diccionario_1993}. The \textbf{Wixarika} training data consists of translated Hans Christian Andersen fairy tales from \citet{mager_probabilistic_2018}.

No formal citation was given for the source of the \textbf{Hñähñu} training data (see \citealp{mager_findings_2021}).

\section{Hyperparameter Details}\label{app:hypers}
\paragraph{Pre-training}
The character embeddings for our multilingual model are initialized by training CBOW \citep{mikolov_efficient_2013} on the AmericasNLP training set for 32 epochs, with a window size of 5. Special tokens like \texttt{<bos>} that do not appear in the training corpus are randomly initialized. These pre-trained embeddings are not frozen during training. 

We pre-train for 16,768 steps, using the Adam optimizer \citep{kingma_adam_2015}. We apply a linear warmup for 1024 steps, and a linear decay afterward. We sweep eight learning rates on a grid of the interval $\left[0.0005, 0.0009\right]$ and encoder dropout values $\{12.5\%, 25\%\}$. A dropout rate of 6.25\% is applied both to the embeddings before being passed to the encoder, and to the hidden-state and start-symbol encodings input to the decoder (see \citealp{downey_masked_2021}). Checkpoints are taken every 128 steps.

\paragraph{K'iche' Transfer Experiments}
Similar to the pre-trained model, character embeddings are initialized using CBOW on the given training set for 32 epochs with a window size of 5, and these embeddings are not frozen during training. 

All models are trained using the Adam optimizer \citep{kingma_adam_2015} for 8192 steps on all but the two smallest sizes, which are trained for 4096 steps. A linear warmup is used for the first 1024 steps (512 for the smallest sets), followed by linear decay. We set the maximum segment length to 10. A dropout rate of 6.25\% is applied to the input embeddings, plus $h$ and the start-symbol for the decoder. Checkpoints are taken every 64 steps for sizes 256 and 512, and every 128 steps for every other size.

For all training set sizes, we sweep 5 learning rates and 3 encoder dropout rates, but the swept set is different for each. For size 256, we sweep learning rates \{5e-5, 7.5e-5, 1e-4, 2.5e-4, 5e-4\} and (encoder) dropout rates \{12.5\%, 25\%, 50\%\}. For size 2048, we sweep learning rates \{1e-4, 2.5e-4, 5e-4, 7.5e-4, 1e-3\} and dropouts \{12.5\%, 25\%, 50\%\}. For the full training size, we sweep learning rates \{1e-4, 2.5e-4, 5e-4, 7.5e-4, 1e-3\} and dropouts \{6.5\%, 12.5\%, 25\%\}.

\begin{sidewaystable*}[ht]
    \centering
    \begin{tabular}{llcccccc}
        \toprule
        Language & File & Lines & Total Tokens & Unique Tokens & Total Characters & Unique Characters & Mean Token Length \\
        \midrule
        All & train.anlp & 259,207 & 2,682,609 & 400,830 & 18,982,453 & 253 & 7.08 \\
        All & train\_balanced.anlp & 171,830 & 1,839,631 & 320,331 & 11,981,011 & 241 & 6.51 \\
        All & train\_downsampled.anlp & 120,145 & 1,284,440 & 255,392 & 8,365,710 & 221 & 6.51 \\
        Asháninka & train.cni & 3,883 & 26,096 & 12,490 & 232,494 & 65 & 8.91 \\
        Asháninka & train\_1.mono.cni & 12,010 & 99,329 & 27,963 & 919,897 & 48 & 9.26 \\
        Asháninka & train\_2.mono.cni & 593 & 4,515 & 2,325 & 42,093 & 41 & 9.32 \\
        Aymara & train.aym & 6,424 & 96,075 & 33,590 & 624,608 & 156 & 6.50 \\
        Bribri & train.bzd & 7,508 & 41,141 & 7,858 & 167,531 & 65 & 4.07 \\
        Guaraní & train.gug & 26,002 & 405,449 & 44,763 & 2,718,442 & 120 & 6.70 \\
        Hñähñu & train.oto & 4,889 & 72,280 & 8,664 & 275,696 & 90 & 3.81 \\
        Nahuatl & train.nah & 16,684 & 351,702 & 53,743 & 1,984,685 & 102 & 5.64 \\
        Quechua & train.quy & 120,145 & 1,158,273 & 145,899 & 9,621,816 & 114 & 8.31 \\
        Quechua & train\_downsampled.quy & 32,768 & 315,295 & 64,148 & 2,620,374 & 95 & 8.31 \\
        Rarámuri & train.tar & 14,720 & 103,745 & 15,691 & 398,898 & 74 & 3.84 \\
        Shipibo Konibo & train.shp & 14,592 & 62,850 & 17,642 & 397,510 & 56 & 6.32 \\
        Shipibo Konibo & train\_1.mono.shp & 22,029 & 205,866 & 29,534 & 1,226,760 & 61 & 5.96 \\
        Shipibo Konibo & train\_2.mono.shp & 780 & 6,424 & 2,618 & 39,894 & 39 & 6.21 \\
        Wixarika & train.hch & 8,948 & 48,864 & 17,357 & 332,129 & 67 & 6.80 \\
        \bottomrule
    \end{tabular}
    \caption{Composition of the AmericasNLP 2021 training sets}
    \label{anlp_train_table}
\end{sidewaystable*}

\begin{sidewaystable*}[ht]
    \centering
    \begin{tabular}{llcccccc}
        \toprule
        Language & File & Lines & Total Tokens & Unique Tokens & Total Characters & Unique Characters & Mean Token Length \\
        \midrule
        All & dev.anlp & 9,122 & 79,901 & 27,597 & 485,179 & 105 & 6.07 \\
        Asháninka & dev.cni & 883 & 6,070 & 3,100 & 53,401 & 63 & 8.80 \\
        Aymara & dev.aym & 996 & 7,080 & 3,908 & 53,852 & 64 & 7.61 \\
        Bribri & dev.bzd & 996 & 12,974 & 2,502 & 50,573 & 73 & 3.90 \\
        Guaraní & dev.gug & 995 & 7,191 & 3,181 & 48,516 & 70 & 6.75 \\
        Hñähñu & dev.oto & 599 & 5,069 & 1,595 & 22,712 & 69 & 4.48 \\
        Nahuatl & dev.nah & 672 & 4,300 & 1,839 & 31,338 & 56 & 7.29 \\
        Quechua & dev.quy & 996 & 7,406 & 3,826 & 58,005 & 62 & 7.83 \\
        Rarámuri & dev.tar & 995 & 10,377 & 2,964 & 55,644 & 48 & 5.36 \\
        Shipibo Konibo & dev.shp & 996 & 9,138 & 3,296 & 54,996 & 65 & 6.02 \\
        Wixarika & dev.hch & 994 & 10,296 & 3,895 & 56,142 & 62 & 5.45 \\
        \bottomrule
    \end{tabular}
    \caption{Composition of the AmericasNLP 2021 validation sets}
    \label{anlp_dev_table}
\end{sidewaystable*}

\end{document}